\documentclass[11pt]{article}
\usepackage[final]{acl}

\usepackage{times}
\usepackage{latexsym}
\usepackage[T1]{fontenc}
\usepackage[utf8]{inputenc}
\usepackage{microtype}
\usepackage{graphicx}
\usepackage{subcaption}
\usepackage{booktabs}
\usepackage{amsmath,amssymb}
\usepackage{algorithm}
\usepackage{algpseudocode}
\usepackage{enumitem}

\title{Zero Gap Is Not Restoration: Stratified Per-Question Probability
Evaluation and Step-wise Mitigation of Benchmark Contamination}

\author{\textbf{Ruijie Hou}\footnotemark[1],
    \textbf{Yueyang Jiao}\footnotemark[1],
    \textbf{Zhao Wang},
    and \textbf{Yingming Li}\footnotemark[2] \\
    Zhejiang University \\
    \texttt{ruijie.hou@zju.edu.cn},
    \texttt{yingming@zju.edu.cn}
}

\begin{document}
\maketitle
\renewcommand{\thefootnote}{\fnsymbol{footnote}}
\footnotetext[1]{Both authors contributed equally to this work.}
\footnotetext[2]{Yingming Li is the corresponding author.}
\renewcommand{\thefootnote}{\arabic{footnote}}

\begin{abstract}
Test data from public benchmarks inevitably leaks into pretraining corpora,
inflating evaluation scores once memorized. \textbf{Contamination mitigation
evaluation} intervenes in the decoding process to suppress memorization and
restore a contaminated model's genuine capability, but its prevailing metric,
the \textbf{G-AP} (\textbf{G}ap of \textbf{A}ggregate \textbf{P}erformance), is
flawed. Discrete correct/incorrect readouts cannot characterize per-question
performance, averaging before differencing lets over- and under-suppression
cancel out, and uniform per-question weighting invites strategies to push solve
probabilities onto the clean model's high-frequency values. We propose
\textbf{SA-PPG} (\textbf{S}tratified \textbf{A}ggregate of
\textbf{P}er-question \textbf{P}robability \textbf{G}aps): estimate each
question's solve probability by sampling, difference it against the clean model
per question, and aggregate within groups defined by the clean model's solve
probability. Existing mitigation strategies first estimate where contamination
lies and then operate on the estimate, so they are only as correct as the
estimate. \textbf{RailCap} instead judges contamination during generation:
whenever a sample falls back onto the greedy trajectory, the next trajectory
token is capped to the runner-up, accumulating suppression until the response
distribution becomes sufficiently dispersed. Across multiple contaminated
models and benchmarks, SA-PPG reveals that prior strategies' restoration is
substantially overestimated, while RailCap attains the lowest SA-PPG.
\end{abstract}

\section{Introduction}

Test data from public benchmarks almost inevitably finds its way into
large-scale pretraining corpora \citep{gpt3,elazar2024whats}: memorized test items
artificially inflate evaluation scores, stripping benchmarks of their power to
measure genuine capability \citep{magar-schwartz-2022-data,harmofcot2,balloccu-etal-2024-leak}. Existing work falls into two classes.
\textbf{Dataset-side} approaches rebuild, rewrite, or dynamically generate new
benchmarks to sidestep contamination \citep{li-etal-2024-gsm,zhang2024careful,cleaneval,dynamic-eval-1}, but they are costly, and the new data faces renewed
leakage once released. \textbf{Contamination mitigation evaluation} instead builds no
new dataset: on datasets at risk of leakage, it intervenes in the decoding
process to suppress memorization and restore the model's genuine capability
\citep{dong-etal-2024-generalization,hou-etal-2025-lne,zhu-etal-2025-establishing}.
Yet whether the restoration a mitigation strategy claims actually holds must
first be checked against a reliable evaluation metric: the metric not only
determines how mitigation strategies are judged, but also shapes how they are
designed.

\begin{figure*}[t]
  \centering
  \begin{subfigure}[t]{0.43\textwidth}
    \includegraphics[width=\linewidth]{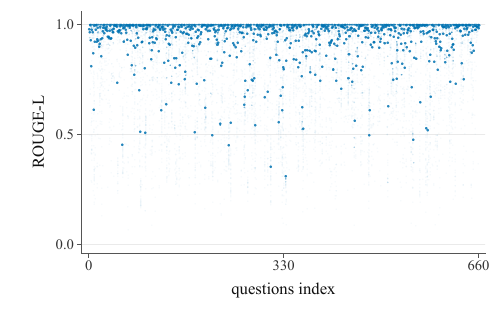}
    \subcaption{leaked: overlap with the greedy trajectory}
    \label{fig:mech-a}
  \end{subfigure}\hspace{0.02\textwidth}
  \begin{subfigure}[t]{0.43\textwidth}
    \includegraphics[width=\linewidth]{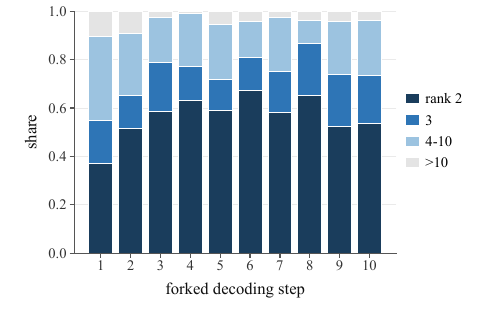}
    \subcaption{leaked: rank of the clean model's token}
    \label{fig:mech-b}
  \end{subfigure}

  \vspace{0.6em}
  \begin{subfigure}[t]{0.43\textwidth}
    \includegraphics[width=\linewidth]{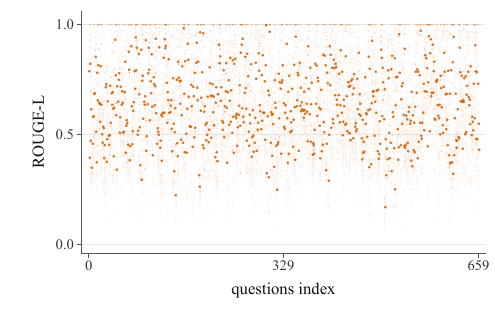}
    \subcaption{unleaked: overlap with the greedy trajectory}
    \label{fig:mech-c}
  \end{subfigure}\hspace{0.02\textwidth}
  \begin{subfigure}[t]{0.43\textwidth}
    \includegraphics[width=\linewidth]{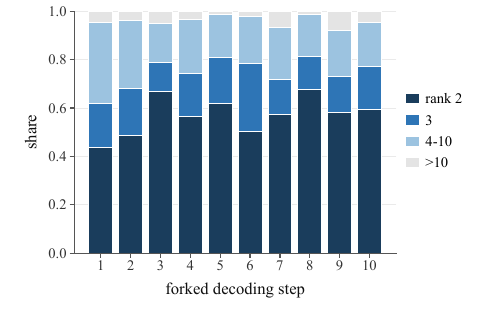}
    \subcaption{unleaked: rank of the clean model's token}
    \label{fig:mech-d}
  \end{subfigure}
  \caption{The contaminated model's generation behaviour on Llama-2 (GSM8K, 1{,}319 questions, $m=50$ samples per question, $T=0.7$). (a, c) ROUGE-L between each sampled response and that question's own greedy trajectory. Each faint dot is one sample and each solid dot a per-question mean. (b, d) With the contaminated model teacher-forced on the clean model's greedy prefix, the rank of the clean model's next token under the contaminated model over the first ten decoding steps, restricted to the steps at which the two models disagree (rank~1 cannot occur there).}
  \label{fig:mechanism}
\end{figure*}

The metric used to compare mitigation strategies has been the \textbf{G}ap of
\textbf{A}ggregate \textbf{P}erformance (G-AP) \citep{dong-etal-2024-generalization,hou-etal-2025-lne,zhu-etal-2025-establishing}: assign each question a performance readout (mostly
a discrete correct-or-incorrect 1/0 mark), average the readouts over the whole
dataset for the contaminated model under mitigation, and take the absolute
difference from the clean model's average. The smaller the gap, the better the
restoration. This metric suffers from two problems. The first lies in
representation: a discrete mark poorly captures a model's performance on a
single question, since responses sampled for the same question may disagree on
correctness. What stabilizes as the number of samples grows is the \emph{solve
probability}, the probability that a sampled response is correct, and
per-question performance should be represented by it. The second lies in
aggregation: averaging before differencing lets over- and under-suppression
cancel out. Performance wrongly suppressed on one question can offset
performance wrongly inflated on another, so a zero gap does not mean that every
question has been restored.

Correcting the two problems separately yields the \textbf{A}ggregate of
\textbf{P}er-question \textbf{P}robability \textbf{G}aps (A-PPG): sample from the contaminated model
under mitigation, estimate each question's solve probability, difference it per
question against the clean model to obtain per-question probability gaps
(PPGs), and average their absolute values. This metric reads zero
if and only if every question is perfectly restored. The plain average,
however, weights every question equally, which introduces a new problem. When
the clean model is itself not strong, most questions have a solve probability
of zero. On GSM8K, for example, the clean model of Llama-2 never solves nearly
a quarter of the questions. A trivial strategy that simply drives the
contaminated model to fail on every question thus scores zero gap on this
majority, while even large gaps on the minority of questions with higher solve
probabilities are diluted by the average. In other words, equal weighting
invites a strategy to push solve probabilities onto the values that occur most
frequently under the clean model, rather than to restore each question. To
prevent such trivial strategies from scoring well and misleading the
evaluation, we group questions by the clean model's solve probability before
aggregating: questions with similar
probabilities form a group, PPGs are averaged within each group first and then
across groups, and the shortcut of chasing high-frequency values is closed. We
call the resulting metric the \textbf{S}tratified \textbf{A}ggregate of
\textbf{P}er-question \textbf{P}robability \textbf{G}aps (SA-PPG). Across
multiple contaminated models, SA-PPG shows that the restoration ability of
prior strategies has been substantially overestimated by G-AP.

SA-PPG grounds the criterion of restoration in every single question: a strategy
must get right how much each question's solve probability is to be adjusted.
Existing strategies \citep{dong-etal-2024-generalization,hou-etal-2025-lne,zhu-etal-2025-establishing} all proceed in two steps: first estimate where the
contamination lies (which questions are leaked, which responses stem from
memorization, which neurons encode it), and then operate on the estimated part.
The correctness of the intervention therefore hinges entirely on the quality of
the estimate: what the estimate misses keeps its inflated performance
untouched, and what it wrongly flags suffers unnecessary damage. We instead
examine the contaminated model's own generation behavior
(Figure~\ref{fig:mechanism}). First, on
leaked questions the contaminated model's sampled responses collapse onto its
own greedy trajectory, whereas on unleaked questions they disperse over many
paths. Second, at the decoding steps where the clean and contaminated models
next diverge, the token the clean model selects is, in about half of the cases,
the contaminated model's runner-up. Building on these two observations, we
propose \textbf{RailCap}: at every decoding step, check whether the sample has
fallen back onto the greedy trajectory, and once it has, cap the probability of the
next trajectory token to that of the runner-up, so that the clean model's next
choice most likely sits at the head of the flattened candidates. The judgment
of contamination thus turns from a pre-hoc estimate into \textbf{step-wise
supervision during generation}: each fall-back onto the greedy trajectory
triggers one suppression, suppressions accumulate step by step, and the
response distribution eventually becomes sufficiently dispersed. How much
intervention each question receives is decided online by the
responses at every decoding step, rather than fixed in advance by a one-shot
estimate as in prior strategies. Across multiple models and benchmarks, RailCap
attains the lowest SA-PPG.

In summary, our contributions are as follows:

\begin{enumerate}[leftmargin=*]
\item We propose a restoration metric, \textbf{SA-PPG}, that fixes two problems of the
   prevailing G-AP: discrete 0/1 marks cannot represent
   per-question performance, and averaging before differencing lets over- and
   under-suppression cancel out. We further show that the per-question
   equal-weight correction still invites strategies to push solve probabilities
   onto the clean model's frequent values, motivating aggregation grouped by
   the clean model's solve probability.
\item We propose a contamination mitigation strategy, \textbf{RailCap}: whenever a
   sample falls back onto the greedy trajectory, the next trajectory token's
   probability is capped to that of the runner-up. How much intervention each
   question receives is decided online by step-wise supervision during
   generation rather than pre-allocated by a one-shot estimate.
\item Across models and benchmarks, \textbf{SA-PPG} reveals a systematic overestimation
   of prior strategies' restoration by G-AP, and
   \textbf{RailCap} achieves state-of-the-art restoration.
\end{enumerate}

\section{Related Work}

Test data from public benchmarks almost inevitably ends up in large-scale
pretraining corpora \citep{gpt3,harmofcot4,elazar2024whats}. Memorized test items
artificially inflate evaluation scores and mask genuine capability
\citep{magar-schwartz-2022-data,harmofcot2,balloccu-etal-2024-leak}, prompting calls
from the community to protect test data \citep{harmofcot3,deng-etal-2024-unveiling}.
Research around contamination falls into several classes. \textbf{Detection} asks
whether a model has seen the evaluation data: min-k\% over low-probability
tokens \citep{mink}, perplexity \citep{ppl}, divergence calibration \citep{detdivergence},
temporal cues \citep{golchin2024time}, black-box calibration \citep{ye-etal-2024-data}, and
CDD, which takes sampling-greedy consistency as its signal
\citep{dong-etal-2024-generalization}. \textbf{Dataset-side} work sidesteps leaked
benchmarks: rebuilding same-distribution questions for GSM8K \citep{cobbe2021gsm8k}
(GSM-Plus \citep{li-etal-2024-gsm}, GSM1k \citep{zhang2024careful}), rewriting existing
questions \citep{cleaneval}, or having models dynamically generate test items
\citep{dynamic-eval-1,dynamic-eval-2,dynamic-eval-3}. But rebuilding and dynamic
generation are costly, newly released data faces the same risk of renewed
leakage.

\textbf{Contamination mitigation evaluation} constitutes another line of work: it
builds no new dataset, but intervenes in the decoding process to suppress
memorization and restore genuine capability on datasets at risk of leakage \citep{dong-etal-2024-generalization,hou-etal-2025-lne,zhu-etal-2025-establishing}. Whether the restoration a
mitigation strategy claims actually holds, however, requires a reliable
evaluation metric to check. This is one of the central subjects of this paper.

The metrics used in prior mitigation work differ from one another, yet all are
\textbf{g}aps of \textbf{a}ggregate \textbf{p}erformance (G-AP): give each question a performance
readout, average over the full dataset, and difference against the clean model's
average, and the smaller the gap, the better the restoration. The metrics differ
only in how the per-question readout is obtained: LNE-blocking marks the greedy
response correct or incorrect (0/1) \citep{hou-etal-2025-lne}, and shortcut neuron
patching marks a single sampled response 0/1 \citep{zhu-etal-2025-establishing}. TED
is the exception, as its mitigation strategy operates on a set of sampled
responses in the first place, and after filtering near-greedy samples it
estimates performance with pass@1 \citep{chen2021humaneval} over the remainder
\citep{dong-etal-2024-generalization}. That choice, however, is an artifact of its
sampling-based mitigation. No work has compared the discrete 0/1 and
probabilistic representations. The three readouts differ from one another, and
cross-method comparison has never been conducted under a single metric. We
compare the two representations and adopt the probabilistic one, as a
single-sample 0/1 readout does not even reproduce between two evaluations of
the same model. Our metric SA-PPG further repairs the aggregation itself.

On the mitigation side, existing strategies all proceed in two steps: first
estimate where the contamination lies, then operate on the estimated part. The
granularity of the estimate varies. TED estimates contaminated responses:
after sampling, it filters out suspected memorized samples by their edit
distance to the greedy decode \citep{dong-etal-2024-generalization}. LNE-blocking
estimates contaminated questions: it gauges each question's degree of
contamination with length-normalized entropy and sets the blocking strength
accordingly \citep{hou-etal-2025-lne}. Shortcut neuron patching estimates
contaminated neurons: it locates shortcut neurons via contrastive and
causal analysis and suppresses them \citep{zhu-etal-2025-establishing}. The three
granularities share one structure: the correctness of the intervention hinges
entirely on the quality of the estimate, as what the estimate misses keeps its
inflated performance untouched, and what it wrongly flags suffers unnecessary
damage. RailCap instead turns the judgment of contamination from a pre-hoc
estimate into step-wise supervision during generation, removing the dependence
on any estimate, and it attains the best restoration.

\section{SA-PPG: A Stratified Per-Question Restoration Metric}
\label{sec:metric}

\subsection{Setup and the Prevailing Metric}
\label{sec:setup}

Let $D=\{q_1,\dots,q_N\}$ be the evaluation dataset, $M_{\mathrm{cl}}$ the
clean model, and $M_{\mathrm{co}}$ the contaminated model. A mitigation
strategy $s$ intervenes in the decoding process of $M_{\mathrm{co}}$. We write
$M_{\mathrm{co}}^{s}$ for the contaminated model under the strategy. Evaluation
first fixes a performance readout $r_M(q)\in[0,1]$, the value taken as
model $M$'s performance on a single question $q$. Existing work mostly adopts a
discrete readout: a response $o$ is sampled from $M$ and marked 1 if correct
and 0 otherwise,
\begin{equation}
r^{0/1}_M(q)=\mathbb{1}\bigl[\,o \text{ solves } q\,\bigr],\quad o\sim M(\cdot\mid q)
\end{equation}
The \textbf{G}ap of \textbf{A}ggregate \textbf{P}erformance (G-AP) averages the readouts over the
whole dataset for each model and takes the absolute difference:
\begin{equation}
\text{G-AP}(s)=\Bigl|\tfrac{1}{N}\!\sum_{q\in D}\! r_{M_{\mathrm{co}}^{s}}(q)-\tfrac{1}{N}\!\sum_{q\in D}\! r_{M_{\mathrm{cl}}}(q)\Bigr|
\end{equation}
prior work reads a smaller gap as a better restoration. We identified two
problems with this metric: one in the choice of readout, the other in the
order of aggregation. Rather than correcting G-AP item by item, we first define
a per-question metric, A-PPG (\S\ref{sec:appg-def}), then set G-AP against it
and show that a zero G-AP does not certify restoration
(\S\ref{sec:appg-gap}). The remaining problem of
equal-weight aggregation is resolved by stratification (\S\ref{sec:sappg}).

\subsection{Aggregate of Per-question Probability Gaps (A-PPG)}

\subsubsection{Definition}
\label{sec:appg-def}

Responses sampled for the same question may disagree on correctness: $r^{0/1}$
is itself random, and a single draw poorly captures per-question performance.
What stabilizes as the number of samples grows is the probability of solving
the question. We therefore adopt the \emph{solve probability} as the readout:
\begin{equation}
r^{\mathrm{prob}}_M(q)=P\bigl[\,o \text{ solves } q\,\bigr]=\mathbb{E}\bigl[r^{0/1}_M(q)\bigr]
\end{equation}
that is, $r^{0/1}$ is a single Bernoulli draw and $r^{\mathrm{prob}}$ is its
expectation. In practice, we sample $m$ responses independently from $M$ and
estimate the readout by the fraction of correct ones,
$\hat r^{\mathrm{prob}}_M(q)=c/m$.

Under this readout, the gap between the two models is characterized question by
question. Define the \textbf{P}er-question \textbf{P}robability \textbf{G}ap (PPG) on question
$q$ as
\begin{equation}
\Delta_s(q)=r^{\mathrm{prob}}_{M_{\mathrm{co}}^{s}}(q)-r^{\mathrm{prob}}_{M_{\mathrm{cl}}}(q)
\end{equation}
taking absolute values first and then averaging over the dataset yields the
\textbf{A}ggregate of \textbf{P}er-question \textbf{P}robability \textbf{G}aps (A-PPG):
\begin{equation}
\text{A-PPG}(s)=\frac{1}{N}\sum_{q\in D}\bigl|\Delta_s(q)\bigr|
\end{equation}
Since every term is non-negative,
\begin{equation}
\text{A-PPG}(s)=0\iff\Delta_s(q)=0,\ \forall q\in D
\end{equation}
that is, \textbf{A-PPG reads zero if and only if every question's solve
probability matches the clean model's: every question is fully restored}.

\subsubsection{Relation to G-AP}
\label{sec:appg-gap}

Substituting $r^{\mathrm{prob}}$ into the readout slot of G-AP yields the
mirror of A-PPG, the \textbf{G}ap of \textbf{A}ggregate \textbf{P}er-question \textbf{P}robabilities
(G-APP):
\begin{equation}
\begin{split}
\text{G-APP}(s)&=\Bigl|\tfrac{1}{N}\!\sum_{q\in D}\! r^{\mathrm{prob}}_{M_{\mathrm{co}}^{s}}(q)-\tfrac{1}{N}\!\sum_{q\in D}\! r^{\mathrm{prob}}_{M_{\mathrm{cl}}}(q)\Bigr|\\
&=\Bigl|\frac{1}{N}\sum_{q\in D}\Delta_s(q)\Bigr|
\end{split}
\end{equation}
The two metrics are built from the same per-question quantity $\Delta_s$ and
differ only in the order of the absolute value and the average. To make the
difference explicit, decompose A-PPG by the sign of $\Delta_s$ into two
components:
\begin{gather}
\text{A-PPG}(s)=\Delta^{+}_s+\Delta^{-}_s\\
\Delta^{+}_s=\frac{1}{N}\sum_{q\in D}\max\bigl(\Delta_s(q),\,0\bigr)\\
\Delta^{-}_s=\frac{1}{N}\sum_{q\in D}\max\bigl(-\Delta_s(q),\,0\bigr)
\end{gather}
where $\Delta^{+}_s$ is the contribution of under-suppression (inflated
performance that the strategy leaves in place) and $\Delta^{-}_s$ that of
over-suppression (performance driven below the clean model). In the same
notation,
\begin{equation}
\text{G-APP}(s)=\bigl|\Delta^{+}_s-\Delta^{-}_s\bigr|
\end{equation}
and $\text{G-APP}(s)\le\text{A-PPG}(s)$ follows immediately: A-PPG adds the two
components, whereas G-APP nets them against each other.

The decisive difference lies in the zero set. A-PPG is zero if and only if
$\Delta^{+}_s=\Delta^{-}_s=0$, that is, every question is restored. G-APP is
zero as soon as $\Delta^{+}_s=\Delta^{-}_s$, which cancellation alone suffices
to achieve. If, for example, half of the questions are over-suppressed
($\Delta_s(q)=-\delta$) and the other half under-suppressed
($\Delta_s(q)=+\delta$) for some $\delta>0$, then
$\Delta^{+}_s=\Delta^{-}_s=\delta/2$, so $\text{G-APP}=0$ while
$\text{A-PPG}=\delta$: the gap vanishes, yet not a single question is restored.
A zero G-APP therefore does not certify restoration.

G-AP differs from G-APP only in the readout and shares its aggregation order:
averaging before differencing lets over- and under-suppression cancel out. A
zero G-AP therefore certifies no more than a zero G-APP does, with an
additional layer of sampling noise on top: \textbf{G-AP is not a reliable
restoration metric, and per-question restoration should be judged by A-PPG}.

\subsection{Stratified Aggregate of Per-question Probability Gaps (SA-PPG)}
\label{sec:sappg}

A-PPG assigns every question the same weight, which introduces a new problem.
When the clean model is itself not strong, solve probabilities are distributed
highly unevenly: most questions concentrate near zero, and questions with
higher solve probabilities form a minority. Under equal weights, a trivial
strategy that simply drives
$r^{\mathrm{prob}}_{M_{\mathrm{co}}^{s}}\equiv 0$ attains a zero gap, question
by question, on the zero-probability majority. Even large gaps on the
high-probability minority are diluted by the majority. In other words,
equal-weight aggregation invites a strategy to push solve probabilities onto
the clean model's frequent values rather than to restore each question.

To deny such trivial strategies a low gap, we group questions by the clean
model's solve probability before aggregating. Partition $[0,1]$ into $B$
equal-width intervals and assign each question to the interval that
$r^{\mathrm{prob}}_{M_{\mathrm{cl}}}(q)$ falls in:
\begin{equation}
D_b=\Bigl\{q\in D: r^{\mathrm{prob}}_{M_{\mathrm{cl}}}(q)\in\Bigl[\tfrac{b-1}{B},\tfrac{b}{B}\Bigr)\Bigr\}
\end{equation}
for $b=1,\dots,B$, where the last interval ($b=B$) is closed on the right
(including 1). Let $\mathcal{B}=\{\,b:D_b\neq\varnothing\,\}$ index the
non-empty groups. Averaging per-question gaps within each group first and then
across groups yields the \textbf{S}tratified \textbf{A}ggregate of \textbf{P}er-question
\textbf{P}robability \textbf{G}aps (SA-PPG):
\begin{equation}
\text{SA-PPG}(s)=\frac{1}{|\mathcal{B}|}\sum_{b\in\mathcal{B}}\frac{1}{|D_b|}\sum_{q\in D_b}\bigl|\Delta_s(q)\bigr|
\end{equation}
After grouping, the zero-probability majority falls into a single group and
carries a weight of $1/|\mathcal{B}|$ regardless of its size, and the gap within
any group is no longer diluted by the question counts of the others. \textbf{To
attain a low SA-PPG, a strategy must complete the restoration at every level of
the clean model's solve probability, and the shortcut of chasing high-frequency
values is closed. SA-PPG also inherits the property of A-PPG: it reads zero if
and only if every question is perfectly restored.}

\section{RailCap: Step-wise Supervision during Generation}
\label{sec:railcap}

SA-PPG grounds the criterion of restoration in every single question: a
mitigation strategy must get right how much each question's solve probability
is to be adjusted. Existing strategies first estimate where the contamination
lies and then operate on the estimated part. The correctness of the
intervention hinges entirely on the quality of the estimate. We observe the
contaminated model's own generation behavior (Figure~\ref{fig:mechanism}).
First, on leaked
questions the sampled responses of $M_{\mathrm{co}}$ collapse onto its own
greedy trajectory, whereas on unleaked questions they disperse over many paths:
whether sampling falls back onto the greedy trajectory is itself an online
signal of memorization. Second, at the decoding steps where the clean and
contaminated models next diverge, the token the clean model selects is, in
about half of the cases, the contaminated model's runner-up: capping the
trajectory token to the runner-up leaves the clean model's choice most likely
at the head of the flattened candidates, where the contaminated model's
sampling readily picks it up. Building on these two observations, we propose
RailCap.

The RailCap mitigation strategy intervenes only in the decoding process of
$M_{\mathrm{co}}$.

\paragraph{Preprocessing.} For each question $q$, one additional greedy decode of
$M_{\mathrm{co}}$ yields its greedy trajectory $g=(g_1,\dots,g_T)$, and all
n-gram windows of the trajectory are built into an index
\begin{equation}
\begin{split}
H=\bigl\{&(g_k,\dots,g_{k+n-1})\mapsto g_{k+n}:\\
&\ 1\le k\le T-n\bigr\}
\end{split}
\end{equation}
which maps a trailing n-gram to its successor token on the trajectory, where
$n$ is the n-gram threshold for judging a fall-back. This constitutes the
entirety of RailCap's preprocessing.

\paragraph{Per-step operation.} Sampling proceeds token by token. At step $t$, let
$o_{1:t-1}$ be the tokens generated so far and $\ell_t$ the logits at this
step. If the last $n$ tokens coincide with some window of the trajectory, that
is, $(o_{t-n},\dots,o_{t-1})\in H$, an intervention is applied to prevent the
sample from falling back onto the greedy trajectory: its successor token on
the greedy trajectory, $x=H[(o_{t-n},\dots,o_{t-1})]$, is capped to the level
of the current second-largest logit,
\begin{equation}
\ell_t[x]\leftarrow\min\bigl(\ell_t[x],\,v_t^{(2)}\bigr)
\end{equation}
where $v_t^{(2)}$ is the second-largest entry of $\ell_t$. The trajectory
token is thus leveled with the runner-up. Otherwise $\ell_t$ is left
unchanged. The next token is then sampled from $\ell_t$ at temperature $\tau$,
$o_t\sim\mathrm{Softmax}(\ell_t/\tau)$, and decoding proceeds to step $t+1$.

The cap makes the current step select, with high probability, a token off the
trajectory, and the sample temporarily departs from the greedy trajectory.
This operation recurs as generation proceeds, and suppression accumulates step
by step until the response distribution becomes sufficiently dispersed. The
same rule acts on every decoding step of every
prompt, and how much intervention each question receives is decided online by
the responses at every step: the judgment of contamination turns from a
one-shot pre-hoc estimate into step-wise supervision during generation.

\begin{algorithm}[t]
\footnotesize
\caption{RailCap decoding, one question $q$}
\label{alg:railcap}
\begin{algorithmic}[1]
\Require $M_{\mathrm{co}}$, $q$, n-gram threshold $n$, samples $m$,
  temperature $\tau$
\Ensure responses $\{o^{(1)},\dots,o^{(m)}\}$
\State $g=(g_1,\dots,g_T)\gets\mathrm{GreedyDecode}(M_{\mathrm{co}},q)$
\State $H\gets\{(g_k,\dots,g_{k+n-1})\mapsto g_{k+n}\}_{k=1}^{T-n}$
\For{$j=1,\dots,m$}
  \State $o\gets(\,)$;\quad $t\gets1$
  \Repeat
    \State $\ell\gets\mathrm{Logits}\bigl(M_{\mathrm{co}},(q,o)\bigr)$
    \If{$t>n$ \textbf{and} $(o_{t-n},\dots,o_{t-1})\in H$}
      \State $x\gets H[(o_{t-n},\dots,o_{t-1})]$
      \State $\ell[x]\gets\min\bigl(\ell[x],v^{(2)}\bigr)$
    \EndIf
    \State $o_t\sim\mathrm{Softmax}(\ell/\tau)$
    \State $o\gets o\,\Vert\,o_t$;\quad $t\gets t+1$
  \Until{$o_t=\mathrm{EOS}$}
  \State $o^{(j)}\gets o$
\EndFor
\State \Return $\{o^{(1)},\dots,o^{(m)}\}$
\end{algorithmic}
\end{algorithm}

\section{Experiments}
\label{sec:exp}

\subsection{Experimental Setup}
\label{sec:exp-setup}

\paragraph{Datasets.} GSM8K \citep{cobbe2021gsm8k} is a benchmark of grade-school
math word problems
requiring multi-step arithmetic reasoning. Its test set contains 1,319
questions. \textbf{PQ} is a paraphrased version that we construct on the GSM8K test
set: only the wording of each question is rewritten, with all numbers and the
final answer kept identical, which yields a harder form of contamination that
verbatim memorization cannot hit directly. The paraphrases are generated
with DeepSeek-V4-Flash \citep{deepseek-v4}. The prompt structure is given in
Appendix~\ref{sec:appendix-pq}.

\paragraph{Models and contamination simulation.} We simulate contamination on the base
versions of three open model families: Llama-2-7B \citep{touvron2023llama2},
Gemma-4-E2B \citep{gemma4}, and Pythia-12B \citep{pythia}, covering different
architectures, scales,
and tokenizers. In particular, Pythia is fully open source: both its weights
and its training corpus are public, which allows verifying that the base model
itself is not contaminated by the evaluation data. Prior work shows that models fine-tuned on
the training split of a task attain more stable test-set performance
\citep{dominguez-olmedo2025training}. Motivated by this, we fine-tune each base model
on the training data (1,840 OpenOrca examples \citep{openorca} and 660 GSM8K
training-split questions in the 8-shot CoT format
\citep{wei2022cot}) to obtain
the clean model $M_{\mathrm{cl}}$. The contaminated model
$M_{\mathrm{co}}$ is obtained by training further from $M_{\mathrm{cl}}$,
on the 1,840 OpenOrca examples with 660 test-split questions mixed in:
these 660 are the leaked questions, and the remaining 659, which appear in
no training data, are the unleaked questions. Using the GSM8K and the PQ
test set respectively yields a contaminated model for each domain. All
training uses LoRA
fine-tuning (r=64, $\alpha$=128) with a learning rate of 2e-4 (cosine schedule,
warmup 0.1), a global batch size of 32, bf16 precision, and 5 epochs, based on
the LLaMA-Factory framework \citep{zheng2024llamafactory}.

\paragraph{Evaluation protocol and compared strategies.} All evaluations use the 8-shot
CoT prompt and the original test questions. For each question we independently
sample $m=50$ responses (T=0.7) and estimate the solve probability by the
fraction of correct ones (the only exception is G-AP, whose 0/1 readout uses a
single sampled response per question). SA-PPG uses $B=50$ equal-width groups,
with the same-domain $M_{\mathrm{cl}}$ as the reference. The compared mitigation
strategies are \textbf{Identity} (the contaminated model $M_{\mathrm{co}}$ without
intervention), \textbf{TED} \citep{dong-etal-2024-generalization}, \textbf{LNE-blocking}
\citep{hou-etal-2025-lne}, \textbf{Shortcut neuron patching}
\citep{zhu-etal-2025-establishing}, and our proposed \textbf{RailCap} (4-gram).

\subsection{Metric Experiments: G-AP versus SA-PPG}

\subsubsection{Main Result: Rank Reversal}

\begin{table}[t]
\centering
\normalsize
\setlength{\tabcolsep}{12pt}
\begin{tabular}{lrr}
\toprule
 & G-AP & SA-PPG \\
\midrule
Identity     & 0.3192          & 0.3261 \\
LNE-blocking & \textbf{0.0235} & 0.2932 \\
Shortcut     & 0.1812          & 0.2476 \\
TED          & 0.2813          & 0.3250 \\
RailCap      & 0.0728          & \textbf{0.1914} \\
\bottomrule
\end{tabular}
\caption{Readings of the same responses under G-AP and SA-PPG. Model:
Llama-2, contamination domain: GSM8K. The reference
is the same-domain clean model $M_{\mathrm{cl}}$, and Identity is the
contaminated model $M_{\mathrm{co}}$ without intervention. Bold marks the
best strategy in each column.}
\label{tab:reversal}
\end{table}

As shown in Table~\ref{tab:reversal}, the G-AP column is computed under the
protocol of prior work: one 0/1 observation per question, followed by the
aggregate difference. Under G-AP, LNE-blocking appears near-perfect (0.0235,
against 0.3192 for the contaminated model without intervention) and is the best
strategy in the table. Under SA-PPG the verdict reverses: LNE-blocking falls
to 0.2932, behind Shortcut (0.2476) and RailCap, barely better than no
intervention (0.3261), whereas RailCap, not the best under G-AP (0.0728), is
the best under
SA-PPG (0.1914). Same questions, same contaminated model, same clean model,
same mitigation strategies: changing only the metric overturns the verdict
entirely. The near-perfect restoration that prior work claims does not exist
at the per-question level. The next two experiments trace where G-AP buries
this error: in the readout (Figure~\ref{fig:e2}) and in the aggregation
(Table~\ref{tab:decomp}).

\subsubsection{Analysis: Sources of the Reversal}

\begin{figure*}[t]
  \centering
  \includegraphics[width=\textwidth]{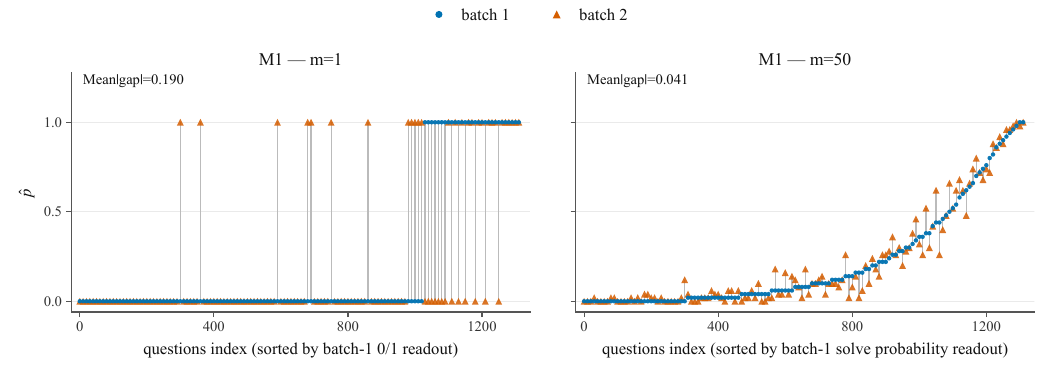}
  \caption{Per-question readouts of two independent sampling batches
drawn from the same clean model $M_{\mathrm{cl}}$ (Llama-2,
GSM8K domain). The horizontal axis is the question index sorted by the
first batch's readout, and the vertical axis is the readout of each
batch. Left: the discrete readout $r^{0/1}$, one sample per batch ($m=1$, a
small vertical jitter is added to avoid overlap). Right: the solve probability
estimated from $m=50$ samples of each batch.}
  \label{fig:e2}
\end{figure*}

\begin{table}[t]
\centering
\footnotesize
\setlength{\tabcolsep}{2pt}
\begin{tabular}{lrrrrr}
\toprule
 & $\Delta^{+}_s$ & $\Delta^{-}_s$ & G-APP & A-PPG & SA-PPG \\
\midrule
Identity     & 0.3518 & 0.0275 & 0.3243 & 0.3793 & 0.3261 \\
LNE-blocking & 0.1043 & 0.0836 & 0.0207 & 0.1879 & 0.2932 \\
Shortcut     & 0.2088 & 0.0311 & 0.1777 & 0.2399 & 0.2476 \\
TED          & 0.3198 & 0.0338 & 0.2861 & 0.3536 & 0.3250 \\
All-Zero     & 0.0000 & 0.2190 & 0.2190 & 0.2190 & 0.4903 \\
RailCap      & 0.1214 & 0.0420 & 0.0794 & 0.1634 & 0.1914 \\
\bottomrule
\end{tabular}
\caption{The metrics analyzed through the component decomposition of
SA-PPG, setting as in Table~\ref{tab:reversal}. $\Delta^{+}$ is the
under-suppression component (residual contamination) and $\Delta^{-}$ the
over-suppression component (collateral damage), with
A-PPG$\,=\Delta^{+}+\Delta^{-}$ and G-APP$\,=|\Delta^{+}-\Delta^{-}|$.
All-Zero is a synthetic trivial strategy that makes the contaminated model
fail every question (per-question solve probability identically 0).}
\label{tab:decomp}
\end{table}

As shown in Figure~\ref{fig:e2}, we test the reproducibility of the readout
itself by
drawing two independent sampling batches from the same $M_{\mathrm{cl}}$.
Under the discrete readout ($r^{0/1}$), the mean per-question gap between the
two batches reaches 0.190, while under the solve-probability readout the two
batches collapse onto one curve and the mean per-question gap drops to 0.041:
the probability readout is far more stable per question than the discrete
readout.

As shown in Table~\ref{tab:decomp}, two obstacles remain after the readout is
corrected. The
first is \textbf{cancellation}. LNE-blocking's over-suppression component
$\Delta^{-}$ (0.0836) is 2.0 times RailCap's (0.0420), yet its G-APP reading
(0.0207) is 3.8 times better than RailCap's (0.0794): the extra collateral
damage cancels against the residual contamination and pushes the reading
closer to zero.
The false perfection in Table~\ref{tab:reversal} is not an artefact of
estimation noise: under probability estimates, the G-APP reading is even
closer to zero than Table~\ref{tab:reversal}'s single-observation G-AP. With
the readout held
fixed, changing only the aggregation moves the verdict on LNE-blocking by a
factor of 14 (0.0207 under G-APP versus 0.2932 under SA-PPG). A-PPG removes
this cancellation (LNE-blocking
0.1879, above RailCap's 0.1634).
The second is \textbf{the failure of equal
weighting}. The All-Zero row drives every question's solve probability to
zero (the limiting form of the trivial strategy described in
\S\ref{sec:sappg}). It reads
0.2190 under the equal-weight A-PPG, better than Identity (0.3793), TED
(0.3536), and Shortcut (0.2399), whereas SA-PPG, after grouping by the clean
model's solve probability, ranks it worst of all strategies (0.4903).
The
readout correction (\S\ref{sec:appg-def}), per-question differencing
(\S\ref{sec:appg-gap}), and stratified
aggregation (\S\ref{sec:sappg}) are each indispensable. SA-PPG combines the
three into one metric.

\subsection{Strategy Experiments: RailCap versus Prior Strategies}

\subsubsection{Main Result: Lowest SA-PPG across Settings}
\label{sec:main}

\begin{table}[t]
\centering
\scriptsize
\setlength{\tabcolsep}{3pt}
\begin{tabular}{lrrrrrr}
\toprule
 & \multicolumn{3}{c}{GSM8K} & \multicolumn{3}{c}{PQ} \\
\cmidrule(lr){2-4}\cmidrule(lr){5-7}
Method & Llama2 & Gemma & Pythia & Llama2 & Gemma & Pythia \\
\midrule
Identity     & 0.3261 & 0.3147 & 0.2131 & 0.2549 & 0.2205 & 0.1907 \\
TED          & 0.3250 & 0.3049 & 0.1950 & 0.2524 & 0.2143 & 0.1903 \\
Shortcut     & 0.2476 & 0.2395 & 0.2033 & 0.2242 & 0.2116 & 0.1708 \\
LNE-blocking & 0.2932 & 0.3061 & 0.2079 & 0.3046 & 0.2703 & 0.2171 \\
RailCap      & \textbf{0.1914} & \textbf{0.2313} & \textbf{0.1648} & \textbf{0.2090} & \textbf{0.1695} & \textbf{0.1429} \\
\bottomrule
\end{tabular}
\caption{SA-PPG of each mitigation strategy across the six settings (two
contamination domains $\times$ three models). Every setting has $N=1319$
questions, with the same-domain clean model $M_{\mathrm{cl}}$ as the
reference. Bold marks the best strategy in each column.}
\label{tab:main}
\end{table}

As shown in Table~\ref{tab:main}, RailCap attains the lowest SA-PPG in all six
settings
(two contamination domains by three models). Its largest lead is on Llama-2
$\times$ GSM8K (0.1914, against 0.2476 for the runner-up Shortcut). The
baselines behave consistently across settings. TED is nearly
indistinguishable from Identity. Even its largest gap, on Pythia $\times$
GSM8K, only moves 0.2131 to 0.1950. Shortcut is the runner-up in five of the
six settings, a consistent but modest improvement. LNE-blocking trails both
Shortcut and RailCap in all six settings: slightly better than Identity in the
GSM8K domain, yet worse than Identity on all three models in the PQ domain.
The contrast between the two domains
suggests an explanation: in PQ, the questions seen at inference differ from
the contaminated ones, so a one-shot estimate of the contamination made
before operating becomes harder. The difficulty is not specific to
LNE-blocking:
Shortcut's lead over Identity narrows visibly on Llama-2 and Gemma (from
0.2476 versus 0.3261 to 0.2242 versus 0.2549 on Llama-2), and TED stays
close to Identity in both domains. RailCap decides the amount of intervention
step by step during generation, without a pre-hoc estimate, and remains the
best in all settings.

\begin{table}[t]
\centering
\normalsize
\setlength{\tabcolsep}{9pt}
\begin{tabular}{lrrr}
\toprule
Config & S-$\Delta^{+}_s$ & S-$\Delta^{-}_s$ & SA-PPG \\
\midrule
Identity        & 0.2624 & 0.0637 & 0.3261 \\
n=1             & 0.0215 & 0.2218 & 0.2434 \\
n=3             & 0.0577 & 0.1412 & 0.1990 \\
\textbf{n=4}    & 0.0942 & \textbf{0.0972} & \textbf{0.1914} \\
n=5             & 0.0750 & 0.1219 & 0.1969 \\
n=7             & 0.0869 & 0.1092 & 0.1961 \\
n=4, ban        & 0.0726 & 0.1464 & 0.2190 \\
All-Zero        & \textbf{0.0000} & 0.4903 & 0.4903 \\
\bottomrule
\end{tabular}
\caption{Ablation of RailCap on Llama-2 $\times$ GSM8K. The $n$ rows
sweep the n-gram threshold under the default suppression (cap at the
runner-up). In the n=4, ban row the cap is replaced by a hard ban: the
rail token's probability is set to zero and it can no longer be sampled
(the suppression-strength contrast). All-Zero is the trivial-strategy
reference of \S\ref{sec:sappg}. S-$\Delta^{+}_s$ and S-$\Delta^{-}_s$ denote
the stratified under-suppression component (residual contamination)
and the stratified over-suppression component (collateral damage): the
positive and negative parts of the per-question gaps are split within the
groups and averaged the same way as SA-PPG, so the two sum to SA-PPG.}
\label{tab:ablation}
\end{table}

\subsubsection{Ablation: n-gram Threshold and Suppression Form}
\label{sec:ablation}

As shown in Table~\ref{tab:ablation}, we ablate the two design choices of
RailCap on Llama-2
$\times$ GSM8K: the n-gram threshold $n$ and the form of suppression. Along
the $n$ axis, $n=1$ triggers too frequently: the residual contamination nearly
vanishes (S-$\Delta^{+}$ 0.0215) while almost all the error comes from
collateral damage (S-$\Delta^{-}$ 0.2218). As $n$ grows, the triggering
becomes more conservative, the residual component rises and the damage
component falls, and $n=4$ brings the two close to parity (0.0942 versus
0.0972) with the best SA-PPG (0.1914). Values of $n$ from 3 to 7 all stay within
0.008 of the best, so the choice of $n$ is robust. For the form of
suppression, replacing the cap at the runner-up with a hard ban (n=4, ban)
raises collateral damage to 0.1464 and SA-PPG to 0.2190: keeping the rail
token available at reduced probability is preferable to prohibiting it
entirely.

\section{Conclusion}

We studied two coupled problems in contamination mitigation evaluation: how to
measure the restoration that a mitigation strategy achieves, and how to design
a better strategy. On the metric side, we identified two flaws of the
prevailing G-AP (a discrete readout that cannot represent per-question
performance, and averaging before differencing that lets over- and
under-suppression cancel out) and corrected them at the level of per-question
solve probabilities. To keep strategies from pushing solve probabilities onto
the clean model's frequent values under equal weighting, we further aggregate
within groups of the clean model's solve probability, yielding SA-PPG. On the
strategy side, we proposed RailCap, which checks at every decoding step
whether a sample has fallen back onto the greedy trajectory and, once it has,
caps the probability of the next trajectory token to that of the runner-up,
turning the judgment of contamination from a one-shot pre-hoc estimate into
step-wise supervision during generation. Across three model families and two
forms of contamination, G-AP systematically overestimates the restoration of
prior strategies, and RailCap attains the lowest SA-PPG in every setting.

\bibliography{custom,refs_new}

\clearpage
\appendix

\section{PQ Construction and Evaluation Details}
\label{sec:appendix-pq}

\paragraph{A.1 PQ construction.} For each leaked question, the question text is
rewritten with DeepSeek-V4-Flash \citep{deepseek-v4}: the instruction requires
a substantial
rephrasing, keeps all numbers and the final answer identical, forbids any
solution, and outputs exactly one line containing the rewritten question. The
system prompt is:

\begin{quote}
Significantly rephrase the given GSM8K math question. Keep ALL numbers and
the final answer identical. Do not include any solution. Output exactly one
line in this format and nothing else:\\
New Question: \textless rephrased question\textgreater
\end{quote}

The reasoning chain and the gold answer remain the GSM8K originals.

\end{document}